\documentclass{article}
\usepackage{ijcai16}
\usepackage{algorithm}
\usepackage{algpseudocode}
\usepackage{pifont}
\usepackage{booktabs}
\usepackage{multirow}
\usepackage{graphicx}
\usepackage{epstopdf}
\usepackage{amssymb,amsthm,bbm,amsmath}

\usepackage{times}

\newtheorem{thm}{Theorem}
\newtheorem{cor}{Corollary}

\newtheorem{lemma}{Lemma}
\theoremstyle{definition}

\theoremstyle{remark}

\newcommand{\bw}{\mathbf{w}}

\newcommand{\bx}{\mathbf{x}}
\newcommand{\by}{\mathbf{y}}

\newcommand{\tr}{\mathbf{T}}
\newcommand{\bX}{\mathbf{X}}
\newcommand{\bXt}{\mathbf{X}_t}

\newcommand{\bB}{\mathbf{B}}
\newcommand{\bbt}{\mathbf{b}_t}
\newcommand{\bb}{\mathbf{b}}
\newcommand{\bP}{\mathbf{P}}

\newcommand{\cB}{\mathcal{B}}
\newcommand{\cD}{\mathcal{D}}

\newcommand{\reals}{\mathbb{R}}

\newcommand{\nr}{\mathbb{R}}

\newcommand{\eqdef}{\triangleq}

\newcommand{\Gcite} [1] {\citeauthor{#1} \shortcite{#1}}
\newcommand{\pecor}{\textbf{CAPE}}
\newcommand{\petcor}{\textbf{CAPE}}
\newcommand{\cape}{\textbf{CAPE}}

\newcommand{\conv}{\text{conv}}

\title{Online Learning of Commission Avoidant Portfolio Ensembles}
\author{Guy Uziel  \\ 
Technion -- Israel Institute of Technology  \\
Ran El-Yaniv  \\ 
Technion -- Israel Institute of Technology  \\
}

\begin{document}

\maketitle

\begin{abstract} 
We present a novel online ensemble learning strategy for portfolio selection.
The new strategy controls and exploits any set of commission-oblivious portfolio selection algorithms.
The strategy handles transaction costs using a novel commission avoidance mechanism. 
We prove a logarithmic regret 
bound for our strategy with respect to optimal mixtures of the base algorithms. Numerical examples 
validate the viability of our method and show significant improvement over the state-of-the-art.
\end{abstract}

\section{Introduction}
Online portfolio selection is a challenging sequential investment problem
introduced by \Gcite{Cover1991}. The problem naturally generalizes
prediction under the logarithmic loss\footnote{From a certain minimax 
perspective portfolio selection isn't any harder than logarithmic loss prediction \cite{CesaL2006}.}
and has become central to online learning research.
One of the major hurdles, abstracted away in the basic formulation (see Section~\ref{sec:ops}),
is \emph{transaction costs}, which can defy many portfolio selection algorithms.
A number of transaction cost aware strategies and regret bounds have been developed,
and the two main techniques used to increase robustness to transaction costs 
were either to reduce such costs by
diluting the number of transaction rounds, as in  
semiconstant-rebalanced portfolios \cite{helmboldSSW1998,kozatS2011},
which rebalance only on a subset of the possible trading days,
or by penalizing (regularizing) costly rebalancing actions  within an online (convex) optimization framework
\cite{DasJB2013,DasJB014}.
All known commission-aware online (adversarial) strategies have been designed to 
track the best (semi) constant rebalanced portfolio (CRP). 
Empirical studies (validated in this paper as well) show that these 
CRP-centric strategies are extremely resilient to very high commission rates.
However, numerous empirical studies also indicate that, without commissions, these 
CRP-driven strategies, and in fact the best CRP in hindsight itself,
achieve inferior performance compared to strategies that are not CRP driven, such as those used by some mean-reversion algorithms
(see, e.g., \Gcite{HuangZLSZ2013,LiH2014}).
To complicate matters, the mean-reverting algorithms, which can achieve phenomenal results without commissions,
are known to be extremely sensitive to commissions (a fact that is also validated here), and some of these 
methods can crash on some datasets even with moderate commissions.

The main question we tackle in this paper is: can we devise a principled method 
that will utilize the better algorithms so as to be more resilient to transaction costs?
We answer this question in the affirmative and
propose an ensemble strategy for controlling and exploiting any set of
commission-oblivious portfolio selection base algorithms.
Our ensemble strategy is
designed to 
track the best convex combination of base algorithms while systematically avoiding 
costly rebalancing activity.


After presenting the strategy and proving a logarithmic regret bound 
with respect to the best in hindsight convex combination (appropriately constrained to reduce commissions),
we present extensive empirical study of our strategy implemented as an ensemble 
over known online portfolio selection algorithms such as 
OLMAR \cite{LiH2012} and Anticor \cite{BorodinETG2004}. 
The strategy can effectively handle a range of  rates (including 1\% proportional transaction costs on almost all the common datasets), and exhibits a graceful performance degradation with commission cost rates.
 Moreover, it consistently outperforms the known commission-aware strategies.

Our learning algorithm, together with its analysis, extend the composite objective mirror descent (COMID) framework of \Gcite{DuchiSST2010}, so as to handle exp-concave loss function (rather than only convex and strongly convex loss functions), which allows our algorithm to achieve $O(\log T)$ regret with respect to the best possible choice in hindsight.
While COMID can be applied as well, it would result in a significantly worse $O(\sqrt{T})$ regret bound.

\section{Online Portfolio Selection }
\label{sec:ops}
In Cover's classic portfolio selection setting 
\cite{Cover1991}, 
we are given a market with $n$ stocks and consider 
an online game between an algorithm and an adversary played through $T$ rounds (say, days).
On each day $t$ the market is represented by
a \emph{market vector} $\bXt$ of relative prices, 
$\bXt \eqdef (x_{1}^{t},x_{2}^{t},...,x_{n}^{t})$,
where for each $i=1, \ldots ,n$, 
$x_{i}^{t}\geq0$ is the \emph{relative price} of stock $i$, defined to be the ratio of its closing price on day $t$ relative to its closing price on day $t-1$. 
We denote by $\bX \eqdef \bX_{1}, \ldots, \bX_{T}$
the sequence of $T$ market vectors for the entire game.
The algorithm's \emph{portfolio} for day $t$ is 
$\bbt \eqdef (b_{1}^{t}, b_{2}^{t} , \ldots, b_{n}^{t})$, where
$b_{i}^{t}\geq0$ is the wealth allocation for stock $i$.
We require that the portfolio satisfy $\sum_{i=1}^{m}b_{i}^{t}=1$. 
Thus, $\bbt$ specifies the online player's 
wealth allocation for each of the $n$ stocks on day $t$, and $b_{i}^t$ is the fraction
of total current wealth invested in stock $i$ on that day.
We denote by $\bB \eqdef  \bb_1, \ldots , \bb_T$ the sequence of
$T$ portfolios played by the algorithm for the entire game. 
The portfolio sequence where all $\bb_i$ equal the same fixed portfolio is called a \emph{constant rebalanced portfolio} (CRP).
At the  start of each trading day $t$, the algorithm
chooses a portfolio $\bb_t$. Thus, by the end of day
$t$, the player's wealth is multiplied by 
$\left\langle \bb_t ,\bXt \right\rangle = \sum_{i=1}^{n}b_i^t x_i^t$,
and assuming initial wealth of \$1, the player's cumulative wealth by the end of the game
is therefore
\begin{equation}
\label{eq:multiplicativeWealth}
R_T (\bB, \bX)  \eqdef  \prod_{t=1}^{T}
\left\langle \bbt ,\bXt \right\rangle.
\end{equation}
In the setting above, it is common to consider the logarithmic cumulative wealth, $\log R_T (\bB, \bX) $, which can be expressed as 
a summation of the logarithmic daily wealth increases, $\log(\left\langle \bbt ,\bXt \right\rangle )$.

 In the online (worst-case) approach to portfolio learning the goal is to online
 generate a sequence $\{ \bb_t \}$ of portfolios that compete with the best-in-hindsight fixed portfolio,
 denoted $\bb_*$. Letting $f_t(\bb)$ be the round $t$ loss of portfolio $\bb$ (in our case, 
 $f_t(\bb) = -\log(\left\langle \bbt ,\bXt \right\rangle )$), we define the regret of sequence 
 $\{ \bb_t \}$ as
 $$ 
 \textbf{Regret} \eqdef \sum^T_{t=1} \left(f_t(\bb_t)-f_t(\bb_*)\right) .
 $$
 In this paper we are mainly concerned with portfolio \emph{ensembles}, where the weights $\bb_t$
 are over trading algorithms and $\bb_*$ represents the optimal-in-hindsight fixed ensemble.

%


\subsection{Introducing Transaction Costs}
\label{sec:transactionCost}

The vanilla portfolio selection model presented above abstracts away transaction costs.
These costs should account for several  components including:
\emph{commissions}, \emph{slippage} (a.k.a. \emph{implementation shortfall}), 
and \emph{market impact costs}.
Commissions are service charges required by brokerage firms or exchanges  for 
handling the purchase or sale of securities. 
Slippage  is the price difference between 
the time  we decide to buy/sell a security and the time  the  transaction is actually 
executed in the exchange. 
Market impact costs (which can be related to slippage) are price differentials incurred when supply and demand dynamically respond to our own orders (e.g., a large buy order on a relatively illiquid security is likely to push its price up).

As opposed to modeling brokerage commissions, which follow a fixed and known schedule
(agreed upon with the broker), a precise modeling of slippage and market impact costs is extremely 
challenging.\footnote{These costs depend on many factors, including, for example, the type of order used, 
liquidity and limit order book dynamics, or  recent transactions history.} 
As a first approximation,  however, it is common to apply a linear transaction cost model
where  each transaction incurs a cost proportional to 
its size 
\cite{BlumK1999,Lobo2007}.\footnote{In practice, market impact costs are often considered to be 
a concave function of the amount traded \cite{Lobo2007}.}
We therefore focus on the following simple multiplicative (proportional) cost model, 
 commonly used in the online portfolio selection literature (see, e.g., \Gcite{Borodin2005}, Sec. 14.5.4). 
In this model, commissions are specified via a fixed parameter, 
$0 < \gamma$, called the \emph{commission rate}, and
for buying (or selling) \$$d$ worth of any stock, the player must pay commission of 
\$$\frac{\gamma}{2} d$.
Thus, the transaction cost incurred when the player rebalances a portfolio 
$\bb$ to portfolio $\bb'$ is $\frac{\gamma}{2} || \bb - \bb' ||_1$.
In the present transaction cost model we assume that commissions are  \emph{self-refinanced}
and the player pays them immediately after performing the daily transactions.
Thus, on day $t$, after rebalancing to portfolio $\bbt$, the market vector  $\bXt$
is revealed and portfolio $\bbt$ becomes
\begin{equation}
\label{eq:bHat}
\hat{\bb}_{t} \eqdef 
\frac{1}{\left\langle \bbt, \bXt \right\rangle }(b_{1}x_{1},b_{2}x_{2}, \ldots, b_{n}x_{n}) .
\end{equation}
Therefore, the commission incurred to rebalance to the next day's portfolio,
$\bb_{t+1}$, is 
\begin{equation}
\label{eq:dailyCosts}
\frac{\gamma}{2} ||  \bb_{t+1} - \hat{\bb}_t ||_1,
\end{equation}
which is paid from the 
current wealth, $\left\langle \bbt,\bXt\right\rangle$.
Altogether, the cumulative wealth of a player paying commission at rate $\gamma$ is
\begin{eqnarray*}
	R_T^{\gamma} (\bB, \bX) =  
	\prod_{t=1}^{T}\left(\left\langle \bbt,\bXt  \right\rangle
	\left[1-\frac{\gamma}{2} ||\mathbf{b}_{t+1}-\hat{\mathbf{b}}_{t}||_{1}\right]\right) .
\end{eqnarray*}

\section{Related Work and Contributions}
\label{sec:Related}

The study of portfolio optimization with transaction costs within mainstream finance 
is a huge topic, beyond our scope. Such studies
typically  
have a traditional operations research flavor where 
stochastic optimization is carried out under specific distributions; see, e.g.,
\Gcite{Davis1990,konno2001,Lobo2007}.
In the brief survey below we only refer to related works  emerging from the online learning (adversarial) line of research initiated by \Gcite{Cover1991}.

\Gcite {BlumK1999} are perhaps the first who studied commissions 
in online portfolio selection and showed an elegant regret analysis for Cover and Ordentlich's 
universal portfolios (UP) algorithm \Gcite{CoverO1996}, which pays proportional commissions.
The idea of semiconstant-rebalanced portfolios (SCRP), which dilute the number of rebalancing trading days for commission reduction, was first mentioned briefly by \Gcite{helmboldSSW1998} and then studied in-depth by  \Gcite{KozatS2008,KozatS2009}, who utilized the context tree weighting
(CTW) 
lossless compression algorithm of \Gcite{Willems1995}
to track the best days for rebalancing. The resulting portfolio algorithm was shown to achieve 
sub-linear regret with respect to the best $k$ days of rebalancing, provided that $k= o(T)$.
Recently, \Gcite{HuangZLZH2015} introduced 
two algorithms, SUP and SUP-$q$, which improve the SCRP algorithms in the sense that
they follow the best (global) CRP (SUP) and best horizon $q$ CRP (SUP-$q$) instead of following 
a specific (given) CRP as SCRP does. The SUP algorithms are shown in \Gcite{HuangZLZH2015} 
to outperform SCRP on many random projections of the NYSE-o and SP500 datasets 
over two stocks.
 

A different approach, presented by \Gcite{DasJB2013,DasJB014}, is called Online Lazy Updates (OLU) and proposes to deal with transaction costs by 
taking $\bb_{t+1}$, the next round portfolio, to be the simplex vector minimizing
\begin{equation*}
\label{eq:Banerjee}
 -\eta\log(\left\langle \bb_{t+1} ,\bXt \right\rangle )+\frac{1}{2}||\bb_{t+1}-\mathbf{b}_{t}||_2^2 + \lambda||\bb_{t+1}-\mathbf{b}_{t}||_1.
\end{equation*}
 The added $\ell_1$ regularization term was introduced to encourage sparse 
 portfolio updates.
 The  idea is to use this  norm as a proxy to the 
true proportional transaction cost incurred by the update
 as given by Equation~(\ref{eq:dailyCosts}).
 A  drawback of this result is that the $O(\sqrt{T})$ regret bound for OLU holds only 
 for $\lambda \approx \frac{1}{\sqrt{T}}$, which means that the effectiveness of this 
 regularizer diminishes as $T$ gets larger.
 
Our motivation for the present work is the empirical observation that the performance  
of the above methods is not consistently satisfying across common benchmark datasets.
While these algorithms can handle commissions very well, their starting point in a setting without commissions is hopeless.
This stems from the fact that they
 are all designed to 
track the performance of
an empirically inferior comparison class, namely the class
of CRP strategies.
As shown
in a number of empirical studies (e.g., \Gcite{BorodinETG2004,LiZHG2012,LiH2012,HuangZLSZ2013}),
even the best CRP \emph{computed in hindsight} itself is not a strong contender relative to other 
known algorithms such as several mean-reversion methods, which attempt to exploit   
recurring statistical inefficiencies in market behavior (e.g., \Gcite{LiH2012}),
and
a family of pattern matching  algorithms proposed by \Gcite{Algoet1988},
\Gcite{GyorfiUW2008,Gyorfi2007}, and \Gcite{LiHG2011}.
See Table~\ref{tab:noComm}, which also validates the disadvantage of the best CRP (BCRP).

Another critical point with regard to the inferiority of CRP-driven methods
is that in a commission-less setting even the most sophisticated universal algorithms 
do not appear to do any better than the simple uniform constant rebalancing portfolio (UCRP).
This observation was first made by \Gcite{BorodinETG2004}. We also observe a similar phenomenon 
in a setting with commissions, where UCRP achieves comparable performance to the known commission aware 
(CRP-driven) methods (see Section~\ref{sec:EMPIRICALEVIDENCE}).


In this paper we introduce a novel mechanism for commission avoidance 
combined within a new
learning algorithm for ensembles applied over any set of (commission-oblivious) 
portfolio selection algorithms. We aim to track the best combination of those algorithms rather than
the best CRP.  Our regret analysis for the proposed method yields optimal logarithmic regret.
We report an extensive empirical study of the proposed procedure, 
where we simulate its performance  over the 6 (publicly available) benchmark datasets 
that were used in this area. The results indicate that the new ensemble algorithm outperforms 
all existing methods over a range of commission costs. 

We note also that we succeeded to extend  both the methods
of 
\Gcite{KozatS2008,KozatS2009}
and \Gcite{DasJB2013,DasJB014}
to track algorithms (rather than CRPs) and even devised appropriate regret 
bounds for them (not reported). The empirical results of these extensions were only marginally better 
than the original and we abandoned them.

\section{Commission Avoidant Portfolio Ensemble}

Our \emph{commission avoidant portfolio ensemble} procedure (henceforth, $\pecor$) is constructed over a set of $d$ sub-algorithms, $A_1, \ldots ,A_d $, 
whose portfolios at round $t$ are  represented by the matrix 
$\mathbf{P'_{t}} \eqdef (P_{t,1},...,P_{t,d})$; namely, $P_{t,j}$ is the $n$-ary column vector specifying the round $t$ allocation prescribed by sub-algorithm 
$j = 1, \ldots, d$ to the $n$ stocks.

\begin{figure}[htb!]

	\vskip 0.1in
	\begin{center}
		\centerline{\includegraphics[width=\columnwidth]{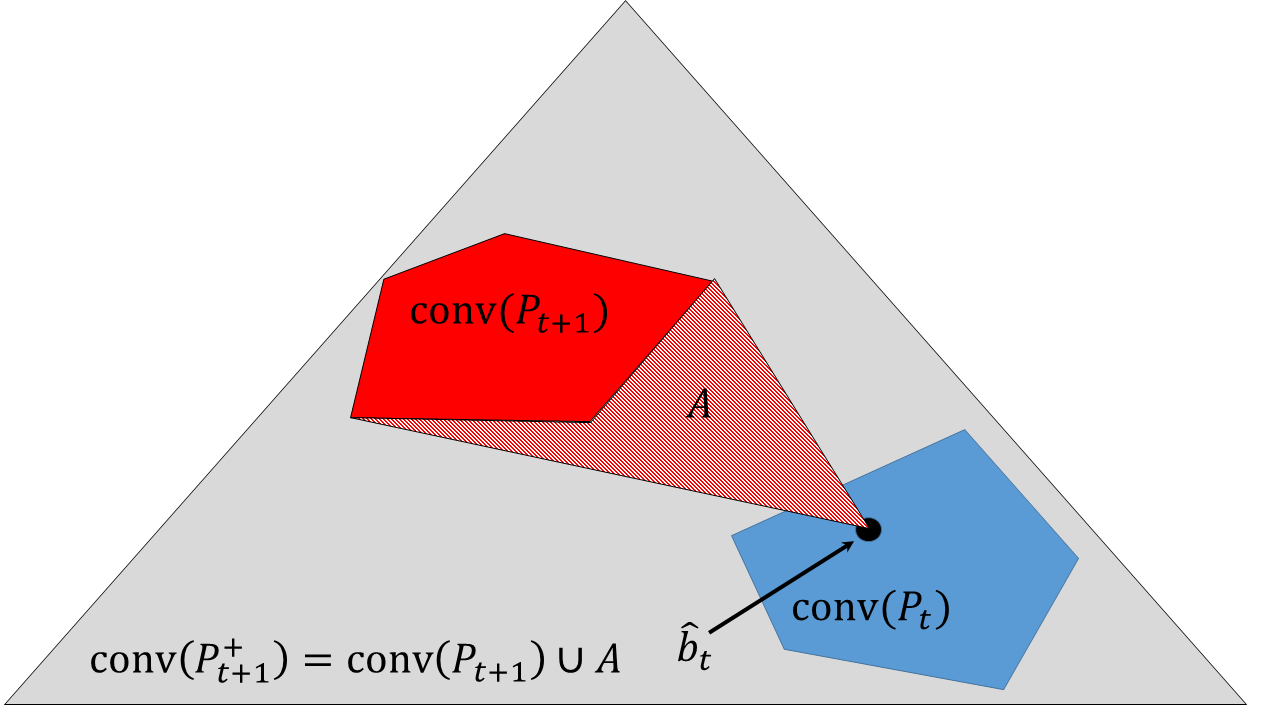}}
		\caption{Inside the stock portfolios simplex (grey triangle) we see the convex hull of round $t$ portfolios of base algorithms (blue pentagon), the convex hull of round $t+1$st portfolios (red pentagon), and the artificial expert  extension (pink area $A$). }
			\label{fig:AE}
	\end{center}
	\vskip -0.1in
\end{figure}

To introduce the new commission avoidance mechanism, we present 
the diagram in Figure~\ref{fig:AE} depicting the probability simplex over $d$ stocks ($d=3$ in the diagram).
Inside the simplex we see $\conv(\bP_t)$, the convex hull of the $d$ portfolios in $\bP_t$ (the blue pentagon,
corresponding to 5 base-algorithms in this diagram).
Inside $\conv(\bP_t)$ we see the current portfolio, $\hat{\bb}_t$, which is depicted inside $\conv(\bP_t)$ but,
of course, may reside outside this pentagon.
The convex hull corresponding to the next round of portfolios, $\conv(\bP_{t+1})$, is also depicted 
(the red pentagon).
Any convex combination of the $d$ sub-algorithms will naturally yield a stock portfolio residing inside 
$\conv(\bP_{t+1})$, and rebalancing to this next portfolio will incur commissions proportional to at least the distance from $\hat{\bb}_t$ to 
$\conv(\bP_{t+1})$.
The idea is to introduce another synthetic expert whose recommended round $t+1$st portfolio 
is precisely $\hat{\bb}_t$. This will extend the next round pentagon to be the union 
of    $\conv(\bP_{t+1})$ and the ``avoidance'' area called $A$ in the diagram (colored in pink).
To this end, we define 
$\bP^+_{t} \eqdef (P_{t,1},...,P_{t,d}, \hat{\bb}_{t-1})$, and therefore,
the above union is simply $\conv(\bP^+_{t+1})$, which  
defines the set of feasible choices for our 
ensemble algorithm for the next round. 
This revised choice allows $\petcor$ to maintain its current holding 
and avoid paying commissions.
$\petcor$ computes in each round $t$ the next allocation vector $\bw_{t+1}$ for each of the
$d$ algorithms and the artificial expert, 
$\bw_{t+1} \eqdef  (w_{1}^{t+1}, \ldots, w_{d+1}^{t+1})$, where the last $d+1$st coordinate is the
weight of the artificial expert.
Denote by $\bw'_{t+1}$ the  projection of $\bw_{t+1}$ over the first $d$ coordinates.
The allocation $\bw_{t+1}$ is optimized using a regularized online Newton step
with the following $\ell_1$ regularizer:
$$
R(\bw'_{t+1})=||\bw^{t+1}_1+...+\bw^{t+1}_d||_1 .
$$
This added penalty encourages $\petcor$ to keep its current holding, thus avoiding 
commissions.
Taking the following $g_t$ to be our loss function,
\begin{equation}
\label{eq:gt}
g_t(\bw) \eqdef -\log (\left\langle \bX_t,\mathbf{P}^+_{t}\bw\right\rangle),
\end{equation}
the regret bound we prove for $\petcor$ in Corollary~\ref{cor:Final} is with respect to $w_*$
which minimizes 
$$
\sum_t g_t(w_* ) + \lambda R(w_*),
$$
for a given $\lambda$.
This optimal static allocation achieves the best possible return with a regularized avoidance of 
rebalancing.
To explain the pseudo-code of $\pecor$ listed in Algorithm~\ref{alg:algo},
we require the following definitions and notation.
Let $A \in \reals^{n \times n} $ be any positive-definite matrix.
For $\bx, \bw \in \reals^n$, the \emph{Bregman divergence} generated by $F_A(\bw) \eqdef \frac{1}{2}\bw^{T}A\bw$ is
$$
\cD_A(\bw||\bx) \eqdef
\frac{1}{2}||\bw-\mathbf{x}||_A^2=\frac{1}{2}(\bw-\mathbf{x})^{T}A(\bw-\mathbf{x}).
$$
We denote by $I_n$  the unit matrix of order $n$.
For a function $f : \reals^{n}\rightarrow \reals$, we denote by $\nabla f(\bw)$ its gradient (if it is differentiable) and by $f'(\bw)$ its subgradient.

\begin{algorithm}[tb]
	\caption{CAPE}
	\label{alg:algo}
	\begin{algorithmic}
		\State {\bfseries Input:} $d$ trading algorithms, \\
		\ \ \ \ \ \ \ \ \ Parameters: $T, \eta,\lambda, \epsilon >0$
		\State {\bfseries Initialize:} $\bP_1^+$, $\bw_1=(\frac{1}{d+1},...,\frac{1}{d+1}), A_{0} = \epsilon I_{d+1}$.
		
		\For{$i=1$ {\bfseries to} $T$}
		\State {\bf Play} $\bw_t$ and suffer loss $g_t(\bw_t)+\lambda R(\bw_t)$
		\State {\bf Compute} portfolios $\bP_{t+1}$ of base algorithms 
		\State {\bf Add} the artificial expert portfolio $\hat{\bb}_{t}$ to get $\bP^+_{t+1}$.
		\State {\bf Update:} $A_{t}=A_{t-1}+\nabla g_{t}(\bw_t)^{\tr}\nabla g_{t}(\bw_t)$ and
		\begin{equation*}
				\mathbf{w}_{t+1}=\mathbf{\arg\min_{w\in \cB}\{}\left\langle \nabla g_{t}(\mathbf{w}_{t}),\mathbf{w}-\mathbf{w}_{t}\right\rangle  +\lambda R(\bw) +\eta \cD_{A_{t}}(\mathbf{w}||\mathbf{w}_{t}))\}
 \end{equation*}
		
		\EndFor
		
	\end{algorithmic}
\end{algorithm}

%
%
%
%
%

In each round $t$,
$\pecor$ first
plays (rebalances its portfolio) according to already computed allocation vector $\bw_t$ (line 4).
In response, the adversary selects a market vector, which determines the 
following loss, $g_t(\bw_t)$, as defined in~(\ref{eq:gt})
(where $\bX_t$ is the market vector selected by the adversary for round $t$).
$\pecor$ then receives $\bP_{t+1} \in \reals^{n \times d}$, the revised portfolios of its sub-algorithms.
Next, $\petcor$ constructs $\bP^+_{t+1}$, the portfolio matrix augmented with the artificial expert, which 
is then used to optimize its next round prediction vector using a regularized online Newton step.
In order to exploit the exp-concavity of the loss function, $\pecor$ utilizes the curvature of the loss function, 
as embedded in the matrix $A_t$, and then uses the Bregman divergence corresponding to $A_t$
so as to optimize its prediction based on second order information (Newton step).

\subsection{Regret Analysis}

Let $\alpha$ be a positive real. A convex function $f$: $\reals^{n}\rightarrow \reals$ is 
$\alpha$\emph{-exp-concave} over the convex domain $\cB \subset \reals^{n}$
if the function $e^{\text{-\ensuremath{\alpha}}f(\mathbf{x})}$ is concave.
It is well known that the class of exp-concave functions strictly contains the class of 
strongly-convex functions. 
For example, the loss function typically used in OPS, 
$f_t(\mathbf{b})=-\log(\left\langle \bb ,\bXt \right\rangle )$,
is exp-concave but not strongly convex.


We conclude this section with two basic lemmas concerning 
exp-concavity that will be used in the proofs 
of Lemma~\ref{lem:10-1}  and Theorem~\ref{thm:Main} that follow.

\begin{lemma}[\Gcite{AK2007}] 
	\label{lem:expConcave}
	Let $f$ be an $\alpha$-exp-concave over $\cB\subset\nr^{n}$
	with diameter $D$, such that $ \forall \bx\in \cB$ , $||\nabla f(\bx)||_{2}\leq G$.
	Then, for $\eta\leq\frac{1}{2}\min\{\alpha,\frac{1}{4GD}\}$, and for
	every $\bx,\by\in \cB$,
	\begin{equation*}
		f(\by)\geq f(\bx)+\left\langle \nabla f(\bx),\by-\bx\right\rangle  
	+\frac{\eta}{2}(\by-\bx)^{T}\left(\nabla f(\bx)\nabla f(\bx)^{T}\right)(\by-\bx) .
	\end{equation*}

\end{lemma}

\begin{lemma}[\Gcite{AK2007}] 
	\label{Exp-concave_hazan}
	Let $f_{t}:\nr^{n}\rightarrow\nr$ be $\alpha$-exp-concave, and  let $A_{t}$ be as in Algorithm~\ref{alg:algo}. 
	Then, for $\eta=\frac{1}{2}\min\{\alpha,\frac{1}{4GD}\}$
	and $\epsilon_0=\frac{1}{\eta^2D^2}$,
	\[\sum_{t=1}^{T}||\nabla f_{t}(\bw_{t})||_{A_{t}^{-1}}^{2}\leq n\log T.\]
\end{lemma}



We consider a standard online convex optimization game \cite{Zinkevich2003} where in 
each round $t$ the online player selects a point $\bw_t$ in a convex set $\cB$; then a convex payoff function $f_t$ is revealed, and the player suffers loss $f_t(\bw_t)$. 
In an adversarial setting, where $f_t$ is selected in the worst possible way,
it is impossible to guarantee absolute online performance. Instead, the objective of the online player is
to achieve sublinear regret relative to the best choice in hindsight, 
$\bw_* \eqdef \arg\min_{w\in \cB} \sum_t f_t(\bw)$, where regret is 
$$
\textbf{Regret} \eqdef \sum^T_{t=1} \left(f_t(\bw_t)-f_t(\bw_*)\right)  .
$$

The mirror descent algorithm 
of \Gcite{NemirovskyY1983} and  \Gcite{BeckT2003} 
for online convex optimization
was extended by \Gcite{DuchiSST2010} as follows.
Instead of solving in each round 
$$
\mathbf{w}_{t+1}=\mathbf{\arg\min_{w\in K}\{}\eta\left\langle \nabla f_{t}(\mathbf{w}_{t}),\mathbf{w}-\mathbf{w}_{t}\right\rangle+\cD(\mathbf{w}||\mathbf{w}_{t}\mbox{)})\},
$$
where $\cD(x||y)$ is the Bregman divergence generated by some strongly convex function $\psi$,
they proposed to solve 
\begin{align*} 
\mathbf{w}_{t+1}&=\mathbf{\arg\min_{w\in K}\{}\eta\left\langle \nabla f_{t}(\mathbf{w}_{t}),\mathbf{w}-\mathbf{w}_{t}\right\rangle +\eta r(\mathbf{w})+\cD(\mathbf{w}||\mathbf{w}_{t}\mbox{)})\} ,
\end{align*}
where $r$ is some convex function which is not necessarily smooth.
They proved that their revised method guarantees $O(\sqrt{T})$ regret relative to 
the best choice in hindsight whenever $f$ is convex.
Moreover, a sharper $O(\log{T})$ regret bound was shown for \emph{strongly convex} $f$. 
This extension, called composite objective mirror descent (COMID),  opened the door to
applications in many fields and, in particular, to the possibility of using an $L_{1}$ regularization term, which encourages sparsity.
In our regret proof we use a similar analysis and extend the COMID framework to deal with exp-concave loss functions.

We assume throughout w.l.o.g. that $r(\bw)\geq 0$, $r(\bw_1)=0$
and that $f_t$ is exp-concave and twice differentiable.



We state without proof the following results (Lemma~\ref{lem:10-1} and Theorem~\ref{thm:Main}). 
Full proofs of these statements will be presented in the long version of 
this paper.

\begin{lemma}
	\label{lem:10-1}
	Let $f_t$ be $\alpha$-exp-concave over $\cB\subset\nr^{n}$
	with diameter $D$, such that $\forall \bw\in \cB$ , $||\nabla f(\bw)||_{2}\leq G$.
	If $\bw_{t}$ is the prediction of Algorithm~\ref{alg:algo} in round $t$,
	then, for $\eta=\frac{1}{2}\min\{\alpha,\frac{1}{4GD}\}$ and for any $\bw_{*}\in \cB$, 
	$$
	\frac{1}{\eta}\left[f_{t}(\bw_{t})-f_{t}(\bw_{*})+r(\bw_{t+1})-r(\bw_{*})\right] \leq
	$$
	$$ 
	\cD_{A_{t-1}}(\bw_{*}||\bw_{t})-\cD_{A_{t}}(\bw_{*}||\bw_{t+1})+\frac{1}{2\eta^{2}}||\nabla f_{t}(\bw_{t})||_{A_{t}^{-1}}^{2} .
	$$
\end{lemma}

\begin{thm}
	\label{thm:Main} 
	Let $f_t$ be $\alpha$-exp-concave over $K\subset\nr^{n}$,
	$\eta=\frac{1}{2}\min\{\alpha,\frac{1}{4GD}\}$ and let $\epsilon_0=\frac{1}{\eta^2D^2}$.
	If $(\bw_{1},\bw_{2}, \ldots ,\bw_{T})$ are the predictions of Algorithm~\ref{alg:algo}, then
	for any $\bw_{*}\in \cB$,
	$$
	\sum_{t=1}^{T}\left(f_{t}(\bw_{t})+r(\bw_{t}\mbox{)}-f_{t}(\bw_{*})-r(\bw_{*}\mbox{)}\right)=O(\log T).
	$$
\end{thm}


\begin{cor}
	\label{cor:Final}
For Algorithm~\ref{alg:algo}, for appropriate\footnote{In our applications we used the parameter $\epsilon =\eta=1$. In general, these parameters can be calibrated according to market variability;
	see  \Gcite{AgarwalHKS2006}.} 
$\epsilon,\eta>0$,
and every $\lambda\geq 0$, for any fixed point $\bw_* \in \cB$, it holds that
	\begin{gather*}
	\sum_{t=1}^{T}g_{t}(\bw_{t})+\lambda R(\mathbf{w})	
	-g_{t}(\bw_{*}) 
	-\lambda R(\mathbf{w}_*) =O(\log T).
	\end{gather*}

\end{cor}

\begin{table*}[htb!]
	\caption{Cumulative wealth of \pecor, its base algorithms and other known commission aware algorithms} \label{Table:Empirical-Results}
	\vskip 0.1in
	\begin{center}
			\begin{small}
				\setlength\tabcolsep{2.5pt}
				\begin{tabular}{l|l|llll|ll|lllll}
					\hline
					
					{\bf Commission rate} &{\bf Dataset}  & \multicolumn{4}{c}{{\bf Base Algorithms}}& \multicolumn{2}{c}{{\bf CAPE}} & \multicolumn{5}{c}{{\bf Other Algorithms}}  \\
					
					& & EG & PAMR & Anticor & OLMAR  & Naive & WF & OLU  & SCRP  & UP & SUP & UCRP \\
					\midrule
					
					\multirow{ 6}{*}{$\gamma=0.25\%$} 
					&NYSE-N          & $28.34$ & $1.67$ & $4.18$E$3$ & $9.8$E$4$ & 
$289$  &$\mathbf{407}$ & $18.06$& 
					$18.94$& $30.70$ & NA & $28.59$\\        
					& NYSE-O         & $25$ & $3.9$E$10$ & $5.8$E$5$ & $3.3$E$11$ & $9.4$E$5$ &$\mathbf{5.4}$E$\mathbf{6}$& $17.49$&$18.86$&  $20.95$ & NA & $24.9$ \\
					&MSCI           & $0.91$ & $0.14$ & $1.73$ & $4.67$ & $1.22$ &$\mathbf{1.4}$& $0.91$ & $0.902$& $0.93$ & $0.66$ & $0.91$ \\
					&DJIA           & $0.8$ & $0.2$ & $1.28$ & $1.47$ & $\mathbf{1.12}$ & $1.03$ & $0.824$ &$ 0.77$ &$0.82$ & $0.70$   &$ 0.78$\\
					&TSE            & $1.55$ & $23.6$ & $13.52$ & $19.25$ & $\mathbf{9.85}$ & $7.84$ & $$1.62$$& $1.618$&  $1.46$ & $1.63$ & $1.55$\\ 
					&SP500            & $1.59$ & $0.3$ & $3.08$ & $1.32$ & 
$2.09$ &$\mathbf{2.31}$ & $1.35$ & $1.419$ &  $1.60$ & $1.43$ & $1.60$\\ 
					\midrule
					\multirow{ 6}{*}{$\gamma=0.5\%$} 
					&NYSE-N          & $25.73$ & $0$ & $82.74$ & $170.1$ & $126.8$ & $\mathbf{144.4}$  &  $18.02$&  $18.8$& $29.35$ & NA & $25.9$ \\
					&NYSE-O            & $23.08$ & $1.9$E$5$ & $1.69$E$4$ &  $6.3$E$8$ & $\mathbf{1}$E$\mathbf{5}$& $8.6$E$4$ & $17.48$& 
					$18.86$& $19.86$  & NA & $22.9$\\ 
					&MSCI           & $0.9$ & $0.14$ & $1.08$ & $1.35$ & $1.11$ & $\mathbf{1.26}$ &  $0.90$& $0.901$& $0.92$ & $0.56$ & $0.9$ \\
					&DJIA           & $0.79$ & $0.08$ & $1.021$ & $0.88$ & $0.93$ & $\mathbf{1.01}$ & $0.819$  &$0.77$ &$0.82$ & $0.67$ &$ 0.78$\\
					&TSE            & $1.52$ & $2.09$ & $6.36$ & $5.96$ & $\mathbf{6.4}$ & $4.33$& $1.63$& $1.617$& $1.45$ & $1.61$ & $1.52$\\ 
					&SP500            & $1.55$ & $0.02$ & $1.69$ & $0.38$ & $1.79$ & $\mathbf{2.16}$   & $1.33$ & $1.418$ & $1.58$ & $1.39$ & $1.56$ \\ 
					\midrule
					\multirow{ 6}{*}{$\gamma=0.75\%$} 
					&NYSE-N          & $23.44$ & $0$ & $1.62$ & $0.2$ & $55.6$ & $\mathbf{65.66}$&  $18.01 $ & $18.65$ & $28.45$ & NA & $23.4$ \\
					&NYSE-O            & $21.30$ & $1.14$ & $465$ & $1.1$E$6$ & $\mathbf{1.1}$E$\mathbf{4}$ & $4.1$E$3$  &  $17.47$ & $18.86$ & $19.01$ & NA & $21$ \\ 
					&MSCI           & $0.89$ & $0$ & $0.67$ & $0.36$ & $0.75$ &$\mathbf{1.16}$ &  $0.90$ & $0.90$ & $0.92$ & $0.54$  & $0.89$ \\
					&DJIA           & $0.78$ & $0.03$ & $0.8$ & $0.36$ & $0.78$ & $\mathbf{0.98}$   & $ 0.77$ & $0.81$ & $0.59$ &$ 0.67$ &$ 0.78$ \\
					&TSE            & $1.49$ & $0.1$ & $2.99$ & $1.83$ & $\mathbf{4.15}$ &$2.85$  & $1.61$ & $1.617$ & $1.43$ & $1.58$ & $1.48$ \\ 
					&SP500            & $1.51$ & $0$ & $0.92$ & $0.1$ & $1.52$ & $\mathbf{2.01}$  & $1.33$ & $1.416$ & $1.57$ & $1.35$  & $1.52$ \\ 
					\midrule
					\multirow{ 6}{*}{$\gamma=1\%$} 
					&NYSE-N          & $21.36$ & $0$ & $0.03$ & $0$ & $24.36$  & $\mathbf{36}$ & $17.98$ & $18.6$ & $26.37$ & NA & $21.2$ \\
					&NYSE-O            & $19.66$ & $0$ & $13.09$ &  $2.1$E$3$ & $\mathbf{1.2}$E$\mathbf{3}$   & $440.73$  & $17.4$ & $18.82$ & $18.23$ & NA  & $19.4$\\ 
					&MSCI           & $0.88$ & $0$ & $0.42$ & $0.1$ & $0.75$ & $\mathbf{1.01}$  & $0.90$ & $0.90$ & $0.91$ & $0.44$  & $0.88$ \\
					&DJIA           & $0.77$ & $0.01$ & $0.64$ & $0.1$ & $0.65$ & $\mathbf{0.92}$  & $0.76$ &$ 0.77$ & $0.81$ & $0.55$ &$ 0.78$ \\
					&TSE            & $1.45$ & $0.01$ & $1.41$ & $0.56$ & $\mathbf{2.5}$ & $1.69$  & $1.61$ &$1.616$ & $1.43$ & $1.57$ & $1.45$ \\ 
					&SP500            & $1.47$ & $0$ & $0.5$ & $0.03$ & $1.3$ & $\mathbf{1.82}$  & $1.33$ & $1.415$ & $1.52$ &  $1.34$ & $1.48$\\ 
				\end{tabular}
			\end{small}
	\end{center}
		\vskip -0.1in
\end{table*}

\section{Empirical Study}
\label{sec:EMPIRICALEVIDENCE}
In this section we present an empirical study of $\petcor$,
examining how well $\petcor$ 
controls and operates a set of base-algorithms in comparison to 
both the base-algorithms themselves as well as the existing competition.
We selected the following set of base-algorithms,
all of which are implemented in the 
\Gcite{OLPS} OLPS simulator. 
Unless otherwise specified, all critical parameters of the base-algorithms were set to the default parameters of the simulator in all experiments. We selected the following four base-algorithms:
\begin{itemize}
	\item
	Anticor~\cite{BorodinETG2004}: one of the first algorithms 
	designed to exploit mean-reversion via (anti) correlation analysis.
	\item
	Passive Aggressive Mean-Reversion (PAMR)~\cite{LiZHG2012}: designed to exploit mean-reversion using ``passive-aggressive'' learning \cite{CrammerDKSS2006}.
	\item
	Online Moving Average Reversion (OLMAR)~\cite{LiH2012}: 
	designed to exploit mean-reversion based on moving average predictions.  OLMAR is known to be a strong performer in many benchmark datasets.
		\item
		Exponentiated Gradient (EG)~\cite{helmboldSSW1998}: one of the early universal algorithms.
	    This algorithm is CRP-driven and as mentioned above, is not expected to serve as a useful ingerdient 
	    in our ensemble. It is included to validate $\cape$'s ability to avoid its portfolio recommendations.
\end{itemize}
In addition we compare performance to the following 5 algorithms, all discussed in Section~\ref{sec:Related}: 
\begin{itemize}
	\item
	UP:  universal portfolios with commissions of  \Gcite{Cover1991,CoverO1996}. 
	To the best of our knowledge,  this is the first online portfolio selection algorithm that was considered and analyzed with commissions, by \Gcite{BlumK1999}.	
	\item 
	SCRP: semi-constant rebalanced portfolios of \Gcite{kozatS2011}.
	This algorithm is an ensemble of sequences of constant rebalancing portfolios each diluting the number of allowed trading rounds.
	\item
	SUP: An extension of SCRP; instead of following a fixed CRP, it follows BCRP  
	\cite{HuangZLZH2015}.
	\item
	OLU: utilizes gradient steps with an added $\ell_1$ regularization term to encourage ``lazy'' portfolio updates \cite{DasJB2013}. 
		\item
	UCRP: The uniform CRP. This is a fixed uniform rebalancing, which is obviously a naive commission-oblivious 
	 benchmark. 
\end{itemize}

We experimented with the 6 datasets that were used in the relevant literature (and appear in the public domain). 
These datasets span several types of market conditions, number of stocks, and total trading periods. 
It is worth noting that the hardest set to profitably trade (using a long-only portfolio
selection algorithm) among the six is DJIA (the Dow Jones Industrial Average), 
which captures a bear market where 25  of the 30 DJIA stocks declined.  
Some properties of these sets are summarized in Table~\ref{tab:dataset}.

Before presenting the results with commissions, we refer the reader to the interesting performance of 
 the base-algorithms without commissions as summarized in Table~\ref{tab:noComm}.
With the exception of one crash of PAMR in the DJIA set, the three mean-reverting algorithms 
achieve unrealistically outstanding results and clearly outperform EG and BCRP by a wide margin. 
This heavenly success is almost completely eliminated when introducing 
commissions (see below).\footnote{The performance on the NYSE-O dataset (which for years served as the only 
benchmark dataset in this area) remains excellent, for the most part. We note that this dataset is believed 
to suffer from extreme survival bias.}

\begin{table}[t]
	\caption{Some properties of the datasets}
	\label{tab:dataset}
	\vskip 0.15in
	\begin{center}
			\begin{sc}
				\begin{tabular}{lcccc}
					\hline
					
					Dataset & Starting day & $\#$ Days & $\#$ stocks  \\
					\hline
					
					NYSE-N   & 1/1/1983 & 6431 & 23       \\
					NYSE-O   & 7/3/1962 & 5651 & 36       \\
										MSCI    & 4/1/2006 & 1043 & 24 \\

					DJIA    & 1/14/2001 & 507 & 30 \\

					TSE   & 1/4/1994 & 1258 & 88       \\

					SP500 & 1/2/1998 & 1276 & 25\\

					\hline
				\end{tabular}
			\end{sc}
	\end{center}
		\vskip -0.1in
\end{table}

\begin{table}[t]
	\caption{No commissions: known CRP-based and other (mean-reverting) algorithms}
	\label{tab:noComm}
	\vskip 0.15in
	\begin{center}
			\begin{sc}
				\begin{tabular}{l||ll||lll}
					\hline
					
					
					Dataset & EG & BCRP & Anticor & PAMR & OLMAR    \\
					\midrule

					NYSE-N     & $31$  & $119.8$ & $6.2$E$6$ & $1.2$E$6$ & $4$E$8$ \\        
					NYSE-O    & $27.09$  & $250.6$ & $2.4$E$8$ & $5$E$15$ & $6$E$16$   \\
					MSCI     & $0.92$   & $1.5$ & $3.2$ & $15.2$ & $14.8$ \\
					DJIA    & $0.8$   & $1.24$   & $2.29$ & $0.68$  & $2.7$  \\
					TSE    & $1.59$  & $6.78$  & $39.36$ & $264.8$ & $69.9$     \\ 
					SP500  & $1.63$   & $4$  & $5.9$ & $5.1$ & $16.9$    \\ 
					\hline
				\end{tabular}
			\end{sc}
	\end{center}
	\vskip -0.1in
\end{table}

Each dataset was considered with four commission rates 
($\gamma$): 0.25\%, 0.5\%, 0.75\%, and 1.00\%.\footnote{At the time of writing, there are deep discount brokers, such as Interactive Brokers,
	whose maximal rate (all included) is 0.5\%. These high rates will be incurred when trading very cheap stocks. Significantly better rates can be obtained when trading highly priced stocks and 
	through rebates received for 
	large volumes.}
Table~\ref{Table:Empirical-Results} summarizes the results of our experiments,
where cumulative wealth is reported.\footnote{While 
	risk-adjusted measures such as the Sharpe ratio are important as well, the total cumulative wealth has the same scale of the commission paid. In the full version we will also report on other 
	performance measures.}\footnote{NYSE-X results for the SUP algorithm requires huge computational resources and over 72 computation hours were not sufficient to complete these runs.} 
Each row in this table corresponds to a different market and commission rate  ($\gamma$), and columns
correspond to algorithms.
The first block of columns corresponds to the base-algorithms, the second block to $\cape$,
and the last block to the known (commission aware) methods.

Observing first the known methods (third block), we see that all of them are extremely resilient to commissions in this range.
However, their performance is not at all superior to that of the simple uniform-CRP (UCRP), which is among the winners in several datasets. Similar observations were previously made in a setting without commissions where
UCRP achieved near identical results to CRP-driven algorithms such as UP and EG \cite{BorodinETG2004}.  
Next, consider the first block containing the (known) base-algorithms. With the exception of EG
(CRP-driven), all the algorithms crash completely with increasing commission rates on a majority of the datasets.

\begin{figure}[htb!]
	
	\vskip 0.1in
	\begin{center}
		\centerline{\includegraphics[width=\columnwidth]{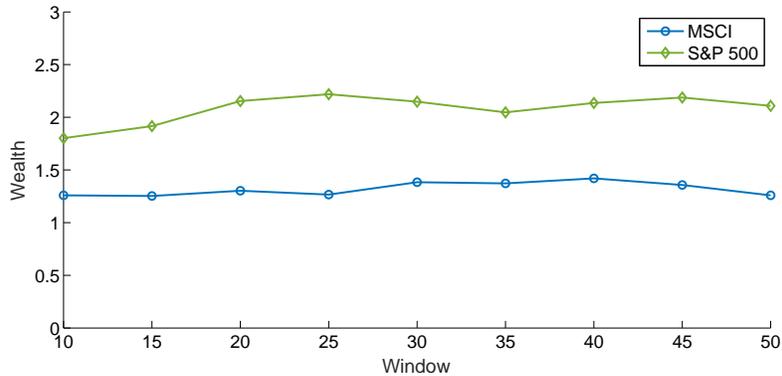}}
		\caption{Wealth sensitivity to window size on two datasets }
		\label{fig:sens}
	\end{center}
	\vskip -0.1in
\end{figure}

For $\cape$ we present two sets of results, based on two different approaches for setting its
hyper-parameter  $\lambda$. The first one, called `Naive', is based on a naive assignment, $\lambda=0.005$, fixed over all datasets and all commission rates.\footnote{This rather arbitrary choice was based on a preliminary examination where we roughly estimated  
the dynamic range of $\lambda$ to be $[0, 0.05]$ over the first dataset.}
Obviously,  one cannot except to find one ``universal'' $\lambda$ that will fit all these 
datasets and commission rates.  Therefore, we also considered a more pragmatic 
approach where we dynamically calibrated the choice of $\lambda$.
 We thus present another set of results for $\cape$ in which this parameter was dynamically optimized using a standard 
walk-forward procedure \cite{Pardo1992},
whereby 
$\lambda$ was sequentially optimized w.r.t. cumulative wealth for the next period over a sliding window of the previous $w$ periods. This setting is called ``WF'' (walk forward).
The WF setting was applied with a fixed $w=25$ for all datasets and commission rates.
To assess the criticality of this choice we conducted a sensitivity analysis where
we computed the sensitivity of total cumulative wealth with respect to choices of 
window size $w \in [10, 50]$ on two entire datasets. The resulting sensitivity graphs are shown 
in Figure~\ref{fig:sens}. It is evident that overall performance is relatively stable as a function of $w$ over these
sets.

Examining first the Naive results (with a fixed $\lambda$ ), we see that $\cape$ outperforms the existing methods
with commission rates 0.25\% and 0.5\% but performance deteriorates with the higher rates, which 
are likely to require stronger regularization (larger $\lambda$).
In the second WF setting (walk-forward) $\cape$ impressively outperforms all competing methods on all datasets.
Moreover, it is evident that $\cape$ successfully exploits the base algorithms even in cases where all of them 
crash. See for example the MSCI dataset with $\gamma = 0.75 \%, 1\%$. 

We examined the portfolio composition of $\cape$ over its base algorithms throughout the runs.
In all datasets and commission rates $\cape$ allocated nearly all its ``non-static'' weight (not allocated 
to the artificial expert) to the two top performing  
base algorithms, which were almost always among the three mean-reverting algorithms (OLMAR, Anticor and
PAMR). The weight allocation for EG was negligible through most of the trading periods. 
This behavior is consistent with the underlying idea behind $\cape$, which is constructed to 
receive its portfolio recommendations from top performing (without commission) base algorithms. 
Indeed, for the most part CRP-driven algorithms such as EG should not be tracked if money is to be made.

\section{Concluding Remarks}
We presented an ensemble learning strategy for portfolio selection algorithms. As far as we know this is the first  commission-aware method designed to exploit any given set of (commission-oblivious) algorithms beyond CRPs in an online adversarial setting.
 Our learning algorithm extends 
 COMID  
 to accommodate  exp-concave functions, and using Newton steps we achieve logarithmic regret bound for our procedure. The demonstrated empirical performance  improves the state-of-the-art across the board,
 both in terms of datasets and commission rate range.

An important challenge would be to combine 
effective mechanisms to dynamically control \emph{risk}.
However, within a \emph{regret minimization} framework, like the one we consider here, 
this challenge is highlighted by the impossibility of achieving 
sub-linear regret in the adversarial setting
with respect to risk adjusted measures such as the Sharpe ratio \cite{EvenKW2006}.


\bibliographystyle{named}
\bibliography{Arxiv_transaction}

\end{document}